\documentclass[runningheads]{llncs}
\usepackage{graphicx}
\begin{document}
\title{Virtual Ground Truth, and Pre-selection of 3D Interest Points for Improved Repeatability Evaluation of 2D Detectors}
\titlerunning{Ground Truth and Informedness}
%
%
\author{Simon R Lang \and
Martin H Luerssen \and
David M Powers}
\authorrunning{S. Lang et al.}
%
\institute{School of Computer Science, Engineering and Mathematics,\\Flinders University, 1284 South Rd, Clovelly Park SA 5042
\email{\{simon.lang,martin.luerssen,david.powers\}@flinders.edu.au}\\
\url{http://www.flinders.edu.au}}
\maketitle              
\begin{abstract}
In Computer Vision, finding simple features is performed using classifiers called interest point (IP) detectors, which are often utilised to track features as the scene changes. For 2D based classifiers it has been intuitive to measure repeated point reliability using 2D metrics given the difficulty to establish ground truth beyond 2D. The aim is to bridge the gap between 2D classifiers and 3D environments, and improve performance analysis of 2D IP classification on 3D objects. This paper builds on existing work with 3D scanned and artificial models to test conventional 2D feature detectors with the assistance of virtualised 3D scenes. Virtual space depth is leveraged in tests to perform pre-selection of closest repeatable points in both 2D and 3D contexts before repeatability is measured. This more reliable ground truth is used to analyse testing configurations with a singular and 12 model dataset across affine transforms in x, y and z rotation, as well as x,y scaling with 9 well known IP detectors. The 
virtual scene's ground truth demonstrates that 3D pre-selection eliminates a large portion of false positives that are normally considered repeated in 2D configurations. The results indicate that 3D virtual environments can provide assistance in comparing the performance of conventional detectors when extending their applications to 3D environments, and can result in better classification of features when testing prospective classifiers' performance. A ROC based informedness measure also highlights tradeoffs in 2D/3D performance compared to conventional repeatability measures.
\end{abstract}
\section{Introduction}
In Computer Vision (CV), the establishment of ground truth so that new feature classification algorithms can be properly measured is an ongoing topic of research. With 3D scanning, printing, and realistic rendering, there are increasing opportunities for CV to be applied to virtual scenes and a multitude of new approaches are exploiting this newly accesible niche \cite{Guo2016}. In the field of 2D CV there are well accepted conventions for measuring interest point/key point based feature detection, the most well known being the work based on research by Schmid and Mikolajczyk \cite{schmid2000} \cite{Mikolajczyk04} that are still regularly used in more recent times \cite{Lindeberg2015}, and have been used a great deal in other CV research relating to interest point/key point repeatability \cite{olag06} \cite{trujillo08} \cite{olague2011}. It is still challenging however to establish a reliable means of establishing better ground truth of real world environments for the purposes of testing 2D based interest 
point detectors \cite{Moreels2005EvaluationOF} \cite{Lindeberg2015} \cite{hansch2014comparison}.

Schmid's metric for evaluation of a set of detectors $K$, classifies points between two pixel arrays $\widetilde{x}_i$, as either repeated, or not and uses a ratio of true positives and true negatives to measure performance. A threshold based on a radial distance $\epsilon$ around each point in the reference scene $\widetilde{x}_1$ determines classification. Equations \ref{eq:schmid1}, \ref{eq:schmid2} and \ref{eq:schmid3} describe this process, with $\widetilde{x}_1$ representing the reference scene as a basis for comparison, and $\widetilde{x}_i$ as the scene image $I_i$ is a member of a set of transforms $j$ being compared. A homography $H_{1i}$ of $\widetilde{x}_i$ enables threshold distances to be measured with $\widetilde{x}_1$, and repeated points to be determined. The default threshold, $\epsilon$=1.5, represents an error rate of 1 pixel distant, also known as the Moore neighborhood, and is considered by Schmid, and researchers in general that apply this metric, to be the optmial tradeoff. Points 
that don't share the same view area are removed from the validation process as they share no valid repeatable point candidates.
\begin{equation}\label{eq:schmid1}
r_{K,J(\epsilon)} = \frac{1}{N-1}\sum_{2}^{N} r_{K,I_i(\epsilon)}\\
\end{equation}
\begin{equation}\label{eq:schmid2}
 r_i(\epsilon)=\frac{|R_i(\epsilon)|}{min(|\widetilde{x}_1|,|\widetilde{x}_i|)}
\end{equation}
\begin{equation}\label{eq:schmid3}
R_i(\epsilon)=\{(\widetilde{x}_1,\widetilde{x}_i) | dist({H_{1i}}, \widetilde{x}_1,\widetilde{x}_i) < \epsilon\}
\end{equation}
\subsection{Repeatability in Virtualised Scenes}
This paper builds on the work done by Lang et al.\cite{Lang:2014:AEI:2675104.2675111} \cite{Lang2013}, where they demonstrated that a virtualised space, whether it be of images, or 3D models, served as a viable testbed for measuring interest point (IP) performance of conventional 2D detectors. Other approaches to IP generation utilise the ground truth of the model directly \cite{Guo2016}, but classifiers that only utilise 2D data are not designed to utilise extra dimensions. This limitation means that they can be highly optimised for 2D scenes, but not 3D, and subsequently also means their performance can't be properly measured in real-world 3D scenarios. Additionally, the lack of ground truth available for optimisation means that 3D applications for 2D based classifiers are constrained.
\section{Methodology}
For the purposes of measuring the performance of IP detectors that utilise only 2D images, a rendering context is utilised to maintain consistency between 2D, and 3D. This preserves 2D consistency of detected IP classification, while also allowing for the precision that the world space of the rendering context provides. Unlike a homography $H_{1i}$ of the pixel positions of points within two scenes $I_1$ and $I_i$, the virtualised scene uses an inverse affine transform $T_{1i}^{-1}$, which enables the precise mapping of detected features to each location in world co-ordinates. Standard Schmmid-based repeatability meausures utilise the pixel positions to determine whether a point is repeated or not, however the pre-selection of points represented as floating point coordinates (a world coordinate system for the 3D rendering context). The pre-selection step is described in equations \ref{eq:preselect_eq1} and \ref{eq:preselect_eq2}, and replaces the algorithm to determine $R_i(\epsilon)$ shown in equation \ref{eq:schmid3}, while not interfering with subsequent processing steps shown in equations \ref{eq:schmid1} and \ref{eq:schmid2}. Additionally all points now include the z worldspace information as described by equation \ref{eq:point}.

To enable 2D/3D pre-selection, $D$ represents the vector dimensions to be utilised when measuring distance, while the function $dist$ determines the distance from the reference point in world space. Pre-selection happens after the removal of points that don't share the same viewport have been removed, but before the points are converted to their pixel positions and $\epsilon$ thresholding is applied. By statically pairing the closest point with its corresponding reference point in 3D space before it is measured in 2D, it enables the comparison of 2D and 3D pre-selection with minimal disruption so that later analysis is simplified.

The testing configuration for 3D pre-selection of points follows the methodology done by \cite{Lang:2014:AEI:2675104.2675111}. It uses a 300x300 image ($I_{i}$) which applies 47 transforms ($J$) of each model in the x and y axis, relative to the viewport as the model is rotated from -50$^{\circ}$ to +50$^{\circ}$ in 10$^{\circ}$ increments (11). The z is rotated from 0$^{\circ}$ to 180$^{\circ}$ in 10$^{\circ}$ increments (19), and the model is scaled in the x,y axis from 1.0, to 4.0 in 0.25 increments (17). This will be applied in two different testing scenarios. The first consisting of a single model, and the other, a dataset of 12 models. Most of the models are 3D scanned, and sourced from commercial, and research sites. The 12 models tested were titled ``bowl'', ``owl'', ``plaque'', ``vase'', ``obelisk'', ``pot''\footnote{http://people.csail.mit.edu/tmertens/textransfer/data/index.html} ``marbles''\footnote{http://www.sci.utah.edu/\textasciitilde wald/animrep/}, ``apple''\footnote{http://www.turbosquid.
com/3d}, ``Stanford bunny'', ``happy Buddha'', ``dragon'' and ``lucy''\footnote{http://graphics.stanford.edu/data/3Dscanrep/}. The ``Stanford asian dragon'' model is tested separately. The bowl, owl, plaque, vase, pot, apple and marbles are textured, and the rest use a generic white mesh.
%
\begin{equation}\label{eq:point}
_i\tilde{x}=(_ix_1, _ix_2, _ix_3)\\
\end{equation}
\begin{equation}\label{eq:preselect_eq1}
R_i(\epsilon)=\{(\widetilde{x}_1,\widetilde{x}_i) |dist(T_{ij} \tilde{x}_1,\tilde{x}_j)\}
\end{equation}
\begin{equation}\label{eq:preselect_eq2}
dist(T_{ij} \tilde{x}_1,\tilde{x}_j)= \sqrt{{\sum_{j=1}} ^D(_1\tilde{x}_j-T^{-1}_{ij}(_ix_j))^2}\\
\end{equation}
The IP detectors tested ($K$) were Harris \cite{harris88}, KLT \cite{klt91}, FAST \cite{rosten_2006_machine,fast06}, SIFT \cite{lowe04} and SURF \cite{herb06} as well as Rohr \cite{rohr92}, Foerstner \cite{foerst86}, Beaudet \cite{beau78} and a different implementation of Harris \cite{harris88}, which have been implemented by the Vigra library \cite{koethe_00_phd-thesis}.
%
%
The process for using pixel-based interest points in a pre-rendered image $I_i$, in conjunction with a world co-ordinate space to determine repeatability is summarised in the following steps.
\label{sec:STEIPR_process}
\textit{\begin{enumerate}
 \item Render scene for each $I_{1i}$ pair with selected affine transform and model
\begin{enumerate}
{\setlength\itemindent{25pt} \item Run detection on 2D image of rendered scenes}
{\setlength\itemindent{25pt} \item Convert 2D interest points to 3D world co-ordinates}
{\setlength\itemindent{25pt} \item Remove 3D points not within model bounding box}
{\setlength\itemindent{25pt} \item Apply inverse affine transform to 3D points where $i>1$}
{\setlength\itemindent{25pt} \item Remove points that don't share overlapping camera views}
{\setlength\itemindent{25pt} \item Pre-select closest distance of $\tilde{x}_i$, to each $\tilde{x}_1$ in world co-ordinates}
\end{enumerate}
\item Repeat 1 for each model
\item Measure repeatability for each model
 \begin{enumerate}
 {\setlength\itemindent{25pt} \item Convert pre-selected closest 3D point pairs to 2D pixel co-ordinate space} 
 {\setlength\itemindent{25pt} \item Calculate repeatability measure of point pairs using corresponding pixel positions to apply $\epsilon$ and mark as repeated or not}
 \end{enumerate}
\item Repeat 3 for each model and aggregate results
\end{enumerate}}
Two different datasets are being used to measure the effects of generalisation, one of these assess repeatability at a more localised level per transform. The asian dragon model was chosen due to its increased non-homogenous surface, protrusions such as horns, and potential for misclassification of repeated points due to lack of depth due to 2D preselection. This affords an analysis based on the effects of generalisation, as well as seeing the effects of preselection for a single model.
\section{Results and Discussion}
\subsection{Analysis of 2D/3D datasets}
%
\begin{figure}[b!]
  \centering
  \includegraphics[width=0.85\textwidth]{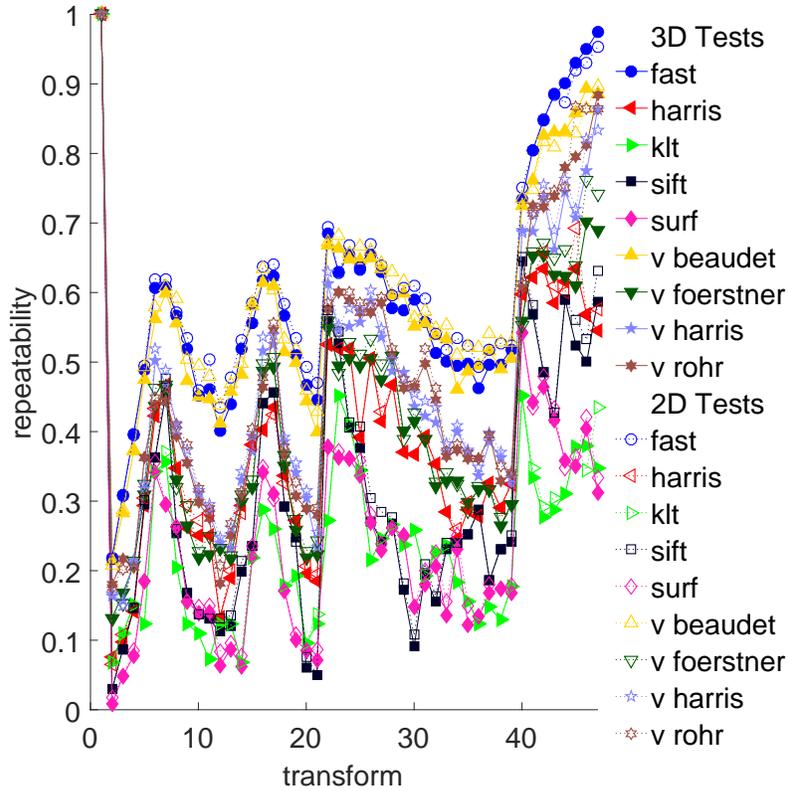}
  \textbf{\caption{Repeatability of asian dragon model at the transform level ($\epsilon$ = 1.5), with rotation in X at 2-11, Y at 12-22, Z at 23-39, and XY scale at 40-47}\label{fig:schmid_all_transforms_dragon2_2D3D}}
\end{figure} 
\begin{figure}
  \centering
  \includegraphics[width=0.75\textwidth,keepaspectratio,trim= 0 0 0 -15]{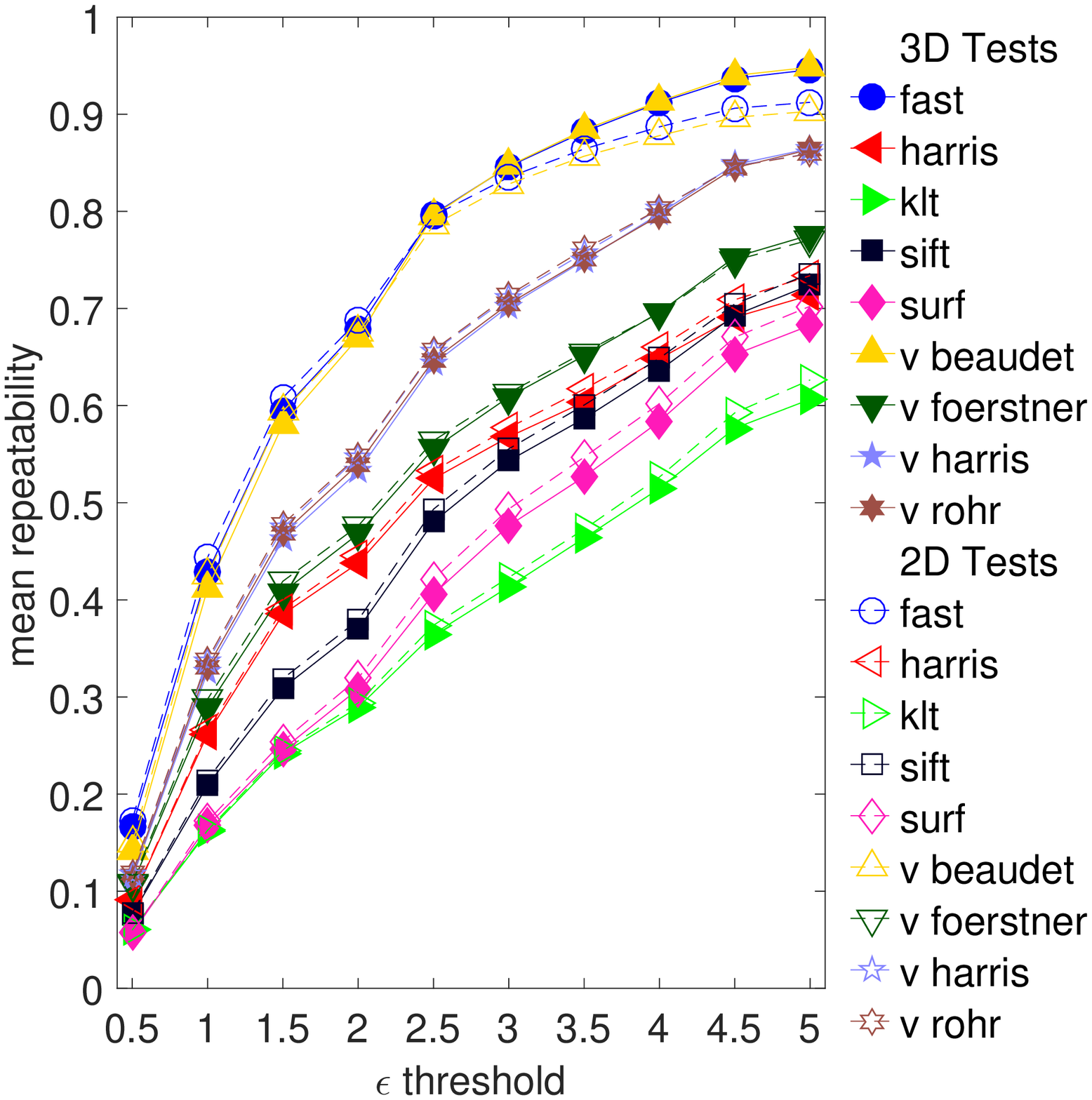}
  \textbf{\caption{Repeatability of asian dragon model}\label{fig:schmid_dragon2_2D3D}}
  \vspace*{\floatsep}
  \includegraphics[width=0.75\textwidth,keepaspectratio,trim= 0 0 0 -15]{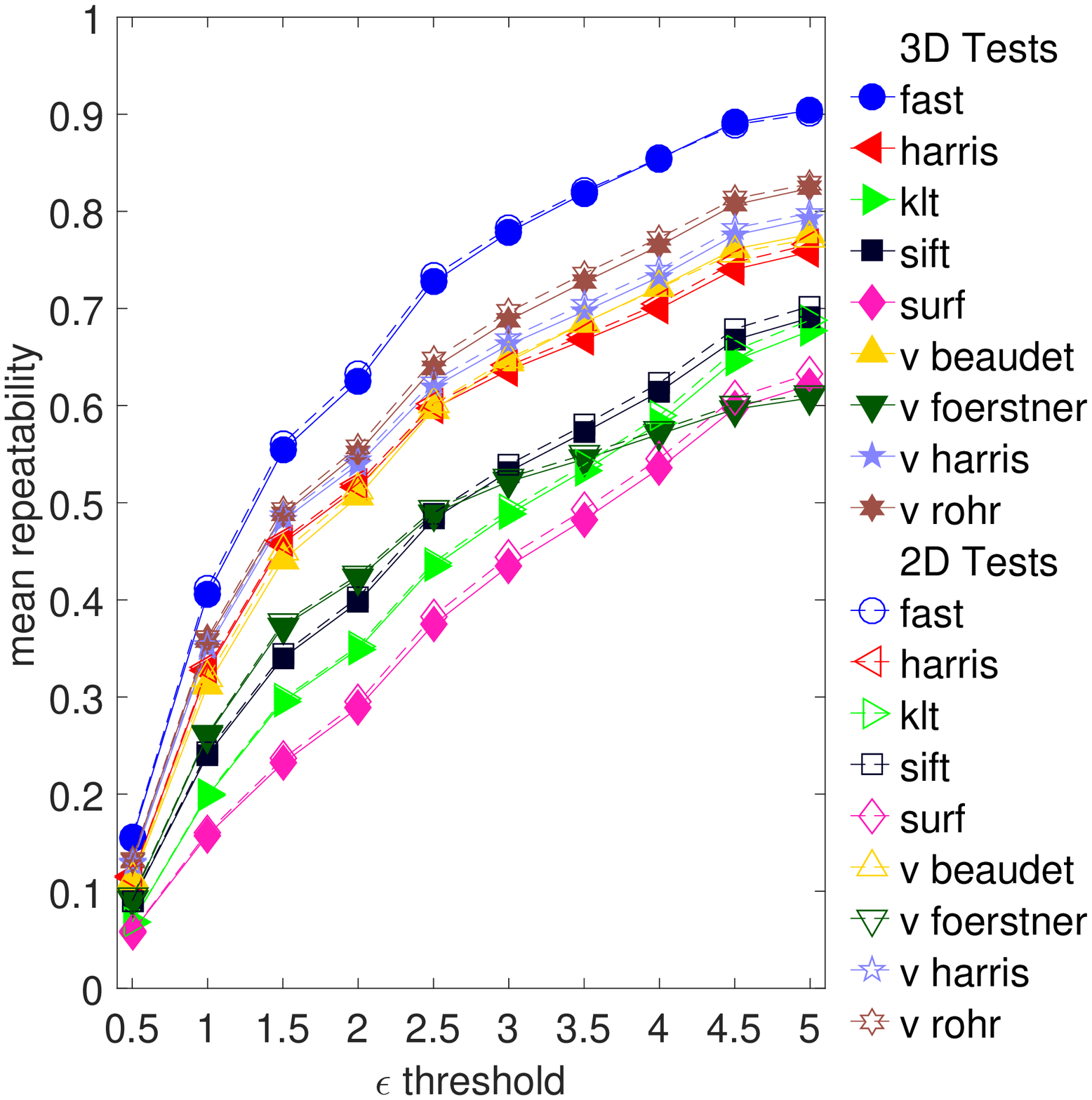}
  \textbf{\caption{Repeatability of 12 model dataset}\label{fig:schmid_12models_2D3D}}
\end{figure}
\begin{figure}
  \centering
  \includegraphics[width=0.68\textwidth,keepaspectratio,trim= 0 0 0 -15]{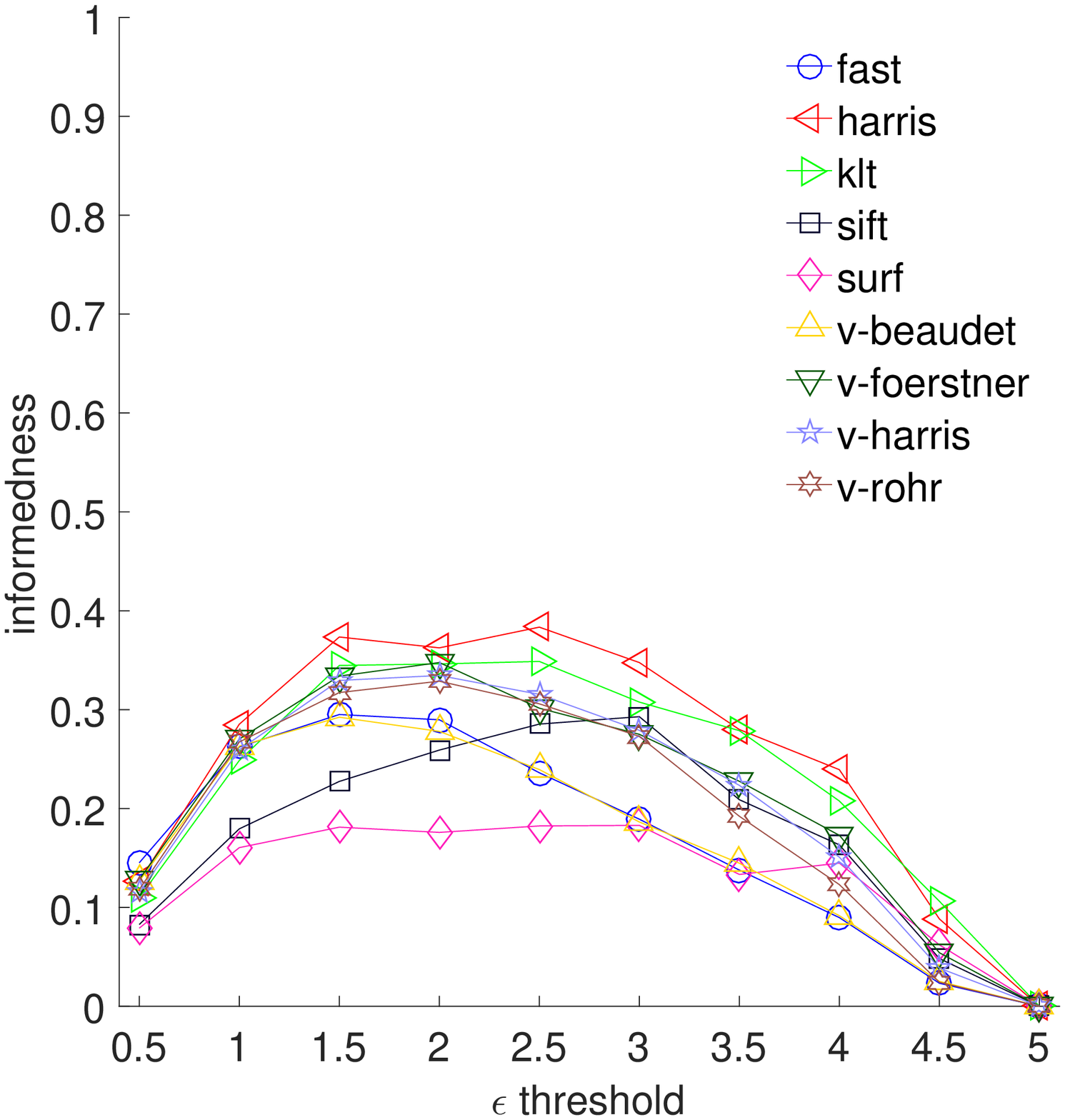}
  \textbf{\caption{Informedness of asian dragon model}\label{fig:inf_dragon2}}
  \vspace*{\floatsep}
  \centering
  \includegraphics[width=0.68\textwidth,keepaspectratio,trim= 0 0 0 -15]{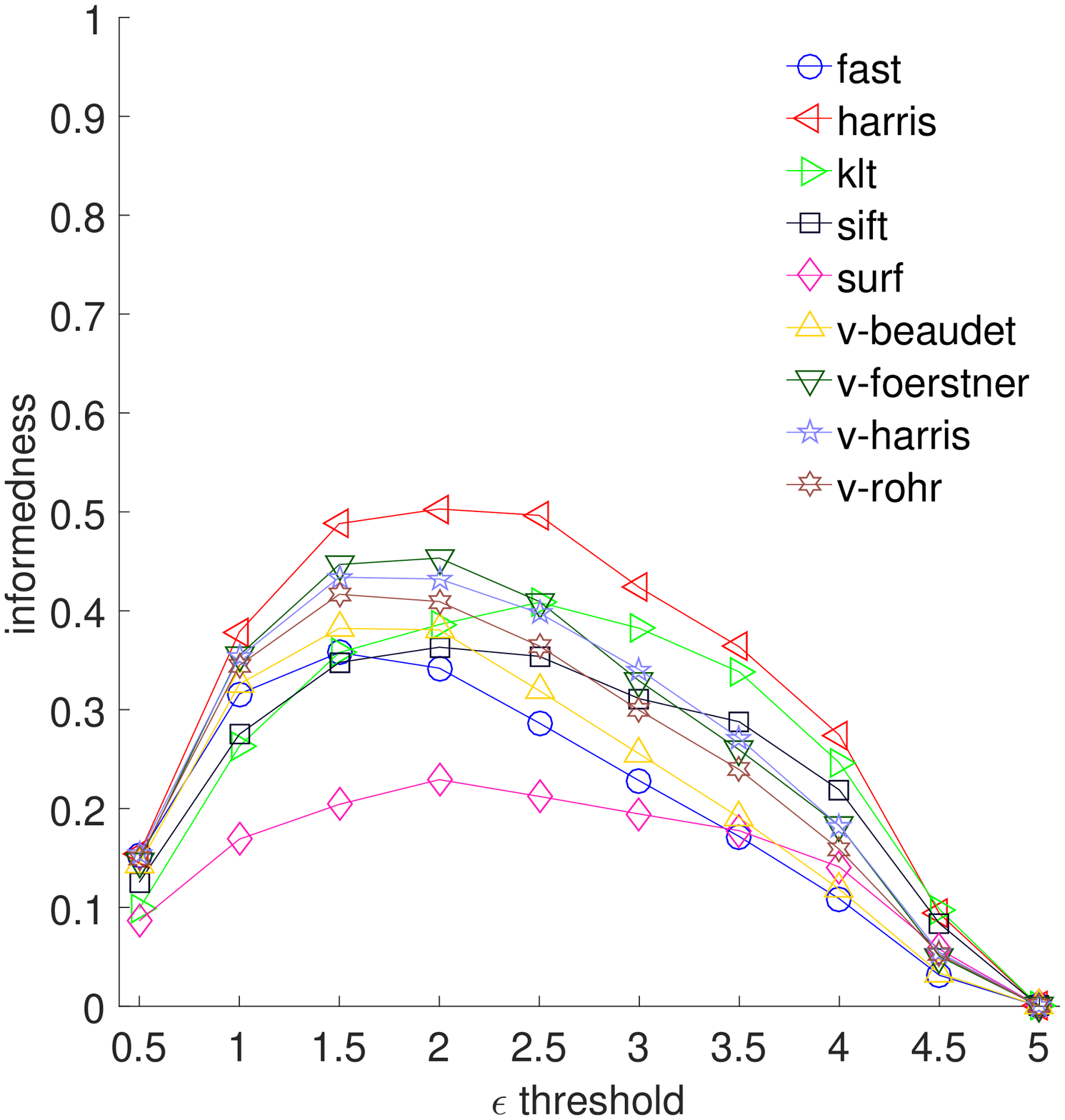}
  \textbf{\caption{Informedness of 12 model dataset}\label{fig:inf_12models}}
\end{figure}
When it comes to comparison of the performance of detectors, the first obvious choice is to compare the repeatability at each epsilon threshold. In most cases an $\epsilon$=1.5 is the preferred threshold for discriminating between detectors. Intuitively, it would be expected that interest points that are able to utilise the depth of the scene would result in more reliable and boosted repeatability rates, given that false positives can be avoided, and better candidates chosen. The results in figure \ref{fig:schmid_dragon2_2D3D} and \ref{fig:schmid_12models_2D3D} highlight that in most cases the 2D preselection of points provides improved repeatability performance, both across detectors, and across most $\epsilon$ thresholds. In many instances more interest points are also detected.

At a superficial level, this could imply that 3D preselection is in fact impacting on performance, and there are indeed a few theoretical corner cases that could justify this. Namely the fact that points could become occluded, and in fact become false positives that are picked up due to them being closer in 2D space compared to other candidate points. This is difficult to justify, however, as there are only a small number of the 47 transforms that could result in this type of occlusion (namely x, and y rotation of the model), and it also would require a very low number of points in order for more unusual or abnormal point candidates to be preselected. Additionally, when examining the asian dragon model at each transform, which can be seen in figure \ref{fig:schmid_all_transforms_dragon2_2D3D}, we can see that repeatability at the scene level shows the same increase for 2D preselection across all scenes. Though it is important to recognise that this is a corner case, the effect (if any) and the criteria 
necessary to exploit this, requires exceptional circumstances. 

To perform a comparison of each dataset that consisted of the singal asian dragon model, and the 12 model dataset, the instances of repeated point pairs between $_1\tilde{x}_j$ and $_i\tilde{x}_j$ for each test were analysed, represented as A and B respectively. To find the $tpr$ we intersect $D_{A}$ and $D_B$ to find true positives common to each testing configuration, and for the $fpr$, intersect and subtract the true positives. This is done at each epsilon which is described in equations \ref{tpr_dsc} and \ref{fpr_dsc}. The intersection of repeated points $D^{3}_{\epsilon}$, which represents the points that utilised 3D data, and $D^{2}_{\epsilon}$ which only used 2D data, provides a ratio of the number of true and false positives within each $\epsilon$ of a point in the reference image, which can be represented as a ROC graph.
\begin{equation}\label{tpr_dsc}
tpr(\epsilon) = \frac{|D^{2}_{\epsilon}\cap D^{3}_{\epsilon}|}{|D^{2}_{R}\cap D^{3}_{R}|}
\end{equation}
\begin{equation}\label{fpr_dsc}
fpr(\epsilon) = \frac{|D^{2}_{\epsilon}-D^{3}_{\epsilon}|}{|D^{2}_{R}-D^{3}_{R}|}
\end{equation}
This provides data sufficient for ROC analysis and calculation of an AUC. However, given the form of analysis that is performed in CV performance, which is to say that it is most common to compare according to the Moore neighborhood ($\epsilon$=1.5), it is difficult to use the data in its current form for comparative analysis. To normalise the $tpr$ and $fpr$ ratios for better comparison, the informedness at each $\epsilon$ threshold can be used, which also provides a performance evaluation that takes into comparison both true positive and false positive detections by each classifier. Informedness is determined by finding the difference between the tpr and fpr and has been demonstrated as being a reliable metric that can determine to a greater extent the similarity of data sets (compared to randomness) \cite{6342144Powers} \cite{DBLP:journals/corr/Powers15a} \cite{DBLP:journals/corr/Powers15b}. The informedness of each detector at each $\epsilon$ threshold can be seen in figure \ref{fig:inf_dragon2} and \ref{fig:inf_12models}.
\subsection{Informedness Optimisation}
Based on the results that are shown in figures \ref{fig:inf_dragon2} and \ref{fig:inf_12models}, there is a clear divergence in the positions of points when compared to preselection of points that finds closest points in a 2D and 3D environment, even though all other testing conditions are identical. It's clear that, unlike figures \ref{fig:schmid_dragon2_2D3D} and \ref{fig:schmid_12models_2D3D} that use only the true positives based on Schmid's approach, there is a substantial mis-classification of points that is not apparent when only true positives are taken into account. This should not be taken as a slight towards true positive repeatability, however, as establishing ground truth is a necessary prerequisite for such an analysis is notoriously hard to reliably or accurately measure in real world environments. It does highlight that there are substantial benefits in adoption of virtualised, or more ideally, 3D scanned real-world objects, so that a more objective ground truth exists that can make these 
performance analyses possible.

Also of note is the fact that in the case of a singular, as well as more generalised dataset, in figures \ref{fig:inf_dragon2} and \ref{fig:inf_12models}, the convention of $\epsilon$=1.5, or Moore neighborhood points, is not necessarily indicative of being the most optimal, especially in the case of detections that are not able to preselect points with the assistance of scene depth. In fact, the informedness data suggests that $\epsilon$=2.0 is generally more favorable across the majority of detectors when tested with the 12 model dataset under the current testing conditions. This informedness of 2D detections indicates that 2.0 should be the more preferred threshold when taking into consideration the tradeoffs of true positives, to false positive detections. Not only does it provide a more rigorous examination of 2D performance compared to 3D, but also indicates at which threshold 2D performance is best, which would be ideal for optimisation when it comes to taking classifiers out of the lab and into the 
real world. These tests demonstrate that the additional metric of informedness, in conjunction with better ground truth testing environments that can effortlessly switch between 2D and 3D, could provide new avenues of performance analysis beyond just concentrating on true positives.
\section{Conclusion}
This paper explores the topic of IP detectors and their repeatability across multiple scene transformations in virtualised 3D spaces with the assistance of 2D and 3D preselection. Though there is a clear move towards utilising 3D ground truth for classifiers that used 3D ground truth natively, 2D classifiers are not able to leverage this benefit. We have sought to formulate a performance analysis that is able to integrate 3D with the assistance of a vitrualised ground truth that gives a more balanced analysis of performance compared to conventional repeatability. It does so by building on the proof of concept that virtualised 3D spaces can be used for testing 2D based IP classifiers, and expands on this by testing the differences between finding nearest neighbor points via 2D and 3D worldspace co-ordinates, by preselecting best candidates before applying traditional repeatability metrics. Testing configurations consisted of a singular model and a 12 model dataset, to compare the effects of gerneralisation, 
and 9 conventional 2D detectors were tested across 47 transforms in x, y and z rotation, and x,y scaling. Though conventional 2D based repeatability showed slightly improved performance, more in depth analysis, made possible due to a more reliable ground truth, highlighted that 2D preselection produced considerable false positives compared to those selected using 3D. This was determined via ROC analysis, and was further refined to a singular performance metric using informedness to normalise results at each $\epsilon$ threshold. Normalisation via informedness also demonstrated that traditional conventional thresholds like only including the Moore neighborhood points as repeatable ($\epsilon$=1.5) are not necessarily optimal, and other thresholds should be considered in 2D contexts for optimisation of classifiers when applied to 2D scenes, in the absence of 3D data.
\section{Future Work}
From the results of our work, there is a substantial difference in the repeatability, and by extension reliability of detected points. We have already begun further research building on this, to measure IP detectors in situations of rapidly prototyping and testing classifiers via Genetic Programming. We aim to build on existing research in CV/GP to explore the effects of virtual ground truth with 3D preselection (and without), to determine its effects on classifier design and performance. Another avenue of research being considered is how effective GP based classifier design is, when taken beyond virtualised 3D spaces into real world environments.

Another area that deserves further exploration is developing a means of preventing occluded points from potentially being preselected. It is common in conventional 3D graphics to use back plane culling where the depth of the scene is used to determine if a face is rendered or not for each pixel. Developing a similar process would help to avoid corner cases. This would be an involved process, and would likely require a sophisticated solution at the shader level, but would be a valuable addition to virtual ground truth environments and IP repeatability such as those where more complex transforms are involved and less points appeared in the scene in question. We consider this an important next step in pursuing interest point evaluation for repeatability purposes.

\begin{thebibliography}{10}
\providecommand{\url}[1]{\texttt{#1}}
\providecommand{\urlprefix}{URL }
\providecommand{\doi}[1]{https://doi.org/#1}

\bibitem{beau78}
Beaudet, P.: Rotationally invariant image operators. In: Proc. Intl. Joint
  Conf. on Pattern Recognition. pp. 579--583 (1978)

\bibitem{fast06}
E.~Rosten, T.D.: Machine learning for high-speed corner detection. In: European
  Conference on Computer Vision (2006)

\bibitem{foerst86}
F{\"o}rstner, W.: A feature based correspondence algorithms for image matching.
  In: Intl. Arch. Photogrammetry and Remote Sensing. vol.~24, pp. 160--166
  (1986)

\bibitem{Guo2016}
Guo, Y., Bennamoun, M., Sohel, F., Lu, M., Wan, J., Kwok, N.M.: A comprehensive
  performance evaluation of 3d local feature descriptors. International Journal
  of Computer Vision  \textbf{116}(1),  66--89 (Jan 2016).
  \doi{10.1007/s11263-015-0824-y}

\bibitem{hansch2014comparison}
H{\"a}nsch, R., Weber, T., Hellwich, O.: Comparison of 3d interest point
  detectors and descriptors for point cloud fusion. ISPRS Annals of the
  Photogrammetry, Remote Sensing and Spatial Information Sciences
  \textbf{2}(3), ~57 (2014)

\bibitem{harris88}
Harris, C., Stephens, M.: A combined corner and edge detector. In: Alvey Vision
  Conference. p. 147–151 (1988)

\bibitem{herb06}
Herbert~Bay, Tinne~Tuytelaars, L.V.G.: Surf: Speeded up robust features. In:
  Lecture Notes in Computer Science. vol.~3951, pp. 404--417 (2006)

\bibitem{klt91}
Kanade, C.T.T.: . detection and tracking of point features. In: Carnegie Mellon
  University Technical Report CMU-CS-91-132 (April 1991)

\bibitem{koethe_00_phd-thesis}
K{\"a}the, U.: Generische Programmierung für die Bildverarbeitung. Ph.D.
  thesis, University of Hamburg (2000)

\bibitem{Lang2013}
Lang, S.R., Luerssen, M.H., Powers, D.M.W.: Proceedings of the 8th
  International Conference on Computer Recognition Systems CORES 2013, chap.
  Repeatability Measurements for 2D Interest Point Detectors on 3D Models, pp.
  361--370. Springer International Publishing, Heidelberg (2013)

\bibitem{Lang:2014:AEI:2675104.2675111}
Lang, S.R., Luerssen, M.H., Powers, D.M.W.: Automated evaluation of interest
  point detectors. Int. J. Softw. Innov.  \textbf{2}(1),  86--105 (Jan 2014).
  \doi{10.4018/ijsi.2014010107}

\bibitem{Lindeberg2015}
Lindeberg, T.: Image matching using generalized scale-space interest points.
  Journal of Mathematical Imaging and Vision  \textbf{52}(1),  3--36 (May
  2015). \doi{10.1007/s10851-014-0541-0}

\bibitem{lowe04}
Lowe, D.G.: Distinctive image features from scale-invariant keypoints.
  International Journal of Computer Vision  \textbf{60},  91--110 (2004)

\bibitem{Mikolajczyk04}
Mikolajczyk, K., Schmid, C.: Scale \&amp; affine invariant interest point
  detectors. International Journal of Computer Vision  \textbf{60},  63--86
  (2004)

\bibitem{Moreels2005EvaluationOF}
Moreels, P., Perona, P.: Evaluation of features detectors and descriptors based
  on 3d objects. Tenth IEEE International Conference on Computer Vision
  (ICCV'05) Volume 1  \textbf{1},  800--807 Vol. 1 (2005)

\bibitem{olague2011}
Olague, G., Trujillo, L.: Evolutionary computer assisted design of image
  operators that detect interest points using genetic programming. Image and
  Vision Computing. Elsevier.  \textbf{29},  484--498 (2011)

\bibitem{olag06}
Olague, G., Trujillo, L.: Using evolution to learn how to perform interest
  point detection. Pattern Recognition, International Conference on
  \textbf{1},  211--214 (2006)

\bibitem{6342144Powers}
Powers, D.M.W.: Roc-concert: Roc-based measurement of consistency and
  certainty. In: 2012 Spring Congress on Engineering and Technology. pp.~1--4
  (May 2012). \doi{10.1109/SCET.2012.6342144}

\bibitem{DBLP:journals/corr/Powers15a}
Powers, D.M.W.: Evaluation evaluation a monte carlo study. CoRR
  \textbf{abs/1504.00854} (2015), \url{http://arxiv.org/abs/1504.00854}

\bibitem{DBLP:journals/corr/Powers15b}
Powers, D.M.W.: Visualization of tradeoff in evaluation: from precision-recall
  {\&} {PN} to lift, {ROC} {\&} {BIRD}. CoRR  \textbf{abs/1505.00401} (2015),
  \url{http://arxiv.org/abs/1505.00401}

\bibitem{rohr92}
Rohr, K.: Modelling and identification of characteristic intensity variations.
  In: Image and Vision Computing. vol.~10, pp. 66--76 (1992)

\bibitem{rosten_2006_machine}
Rosten, E., Drummond, T.: Machine learning for high-speed corner detection. In:
  European Conference on Computer Vision. vol.~1, pp. 430--443 (May 2006)

\bibitem{schmid2000}
Schmid, C., Mohr, R., Bauckhage, C.: Evaluation of interest point detectors.
  International Journal of Computer Vision  \textbf{37},  151--172 (2000).
  \doi{10.1023/A:1008199403446}

\bibitem{trujillo08}
Trujillo, L., Olague, G.: Automated design of image operators that detect
  interest points. Massachusetts Institute of Technology  \textbf{16}(4),
  483--507 (2008)

\end{thebibliography}
\end{document}